\documentclass[sigconf]{acmart}
\AtBeginDocument{%
  }

\setcopyright{acmlicensed}
\copyrightyear{2018}
\acmYear{2018}
\acmDOI{XXXXXXX.XXXXXXX}
\acmConference[Conference acronym 'XX]{Make sure to enter the correct
  conference title from your rights confirmation email}{June 03--05,
  2018}{Woodstock, NY}
\acmISBN{978-1-4503-XXXX-X/2018/06}
\usepackage{diagbox}
\usepackage{pifont}% http://ctan.org/pkg/pifont
\usepackage{dsfont}
\usepackage{soul}
\usepackage{subcaption}
\usepackage{multirow}
\usepackage{tikz}
\usepackage[capitalize]{cleveref}
\Crefname{section}{Section}{Sections}
\Crefname{table}{Table}{Tables}
\Crefname{figure}{Figure}{Figures}
\newcommand{\tikzcircle}[2][red,fill=red]{\tikz[baseline=-0.5ex]\draw[#1,radius=#2] (0,0) circle ;}%
\begin{document}

\title[Interpreting Radiologist's Intention]{Interpreting Radiologist's Intention from Eye Movements \\in Chest X-ray Diagnosis}

\author{Trong Thang Pham}
\affiliation{%
  \institution{University of Arkansas}
  \country{USA}
}

\author{Anh Nguyen}
\affiliation{%
  \institution{University of Liverpool}
  \country{UK}
}

\author{Zhigang Deng}
\affiliation{%
  \institution{University of Houston}
  \country{USA}
}

\author{Carol C. Wu}
\affiliation{%
  \institution{MD Anderson Cancer Center}
  \country{USA}
}

\author{Hien Nguyen}
\affiliation{%
  \institution{University of Houston}
  \country{USA}
}

\author{Ngan Le}
\affiliation{%
  \institution{University of Arkansas}
  \country{USA}
}

\begin{abstract}
Radiologists rely on eye movements to navigate and interpret medical images. A trained radiologist possesses knowledge about the potential diseases that may be present in the images and, when searching, follows a mental checklist to locate them using their gaze. This is a key observation, yet existing models fail to capture the underlying intent behind each fixation.
In this paper, we introduce a deep learning-based approach, \emph{RadGazeIntent}, designed to model this behavior: having an intention to find something and actively searching for it. Our transformer-based architecture processes both the temporal and spatial dimensions of gaze data, transforming fine-grained fixation features into coarse, meaningful representations of diagnostic intent to interpret radiologists' goals.
To capture the nuances of radiologists’ varied intention-driven behaviors, we process existing medical eye-tracking datasets to create three intention-labeled subsets: RadSeq (Systematic Sequential Search), RadExplore (Uncertainty-driven Exploration), and RadHybrid (Hybrid Pattern).
Experimental results demonstrate RadGazeIntent’s ability to predict which findings radiologists are examining at specific moments, outperforming baseline methods across all intention-labeled datasets.
\end{abstract}



\begin{teaserfigure}
    % https://drive.google.com/file/d/15trOQQcH4DC4Xvxw8njSebgpEhl59SKx/view?usp=sharing
    \centering
    \includegraphics[width=0.8\linewidth]{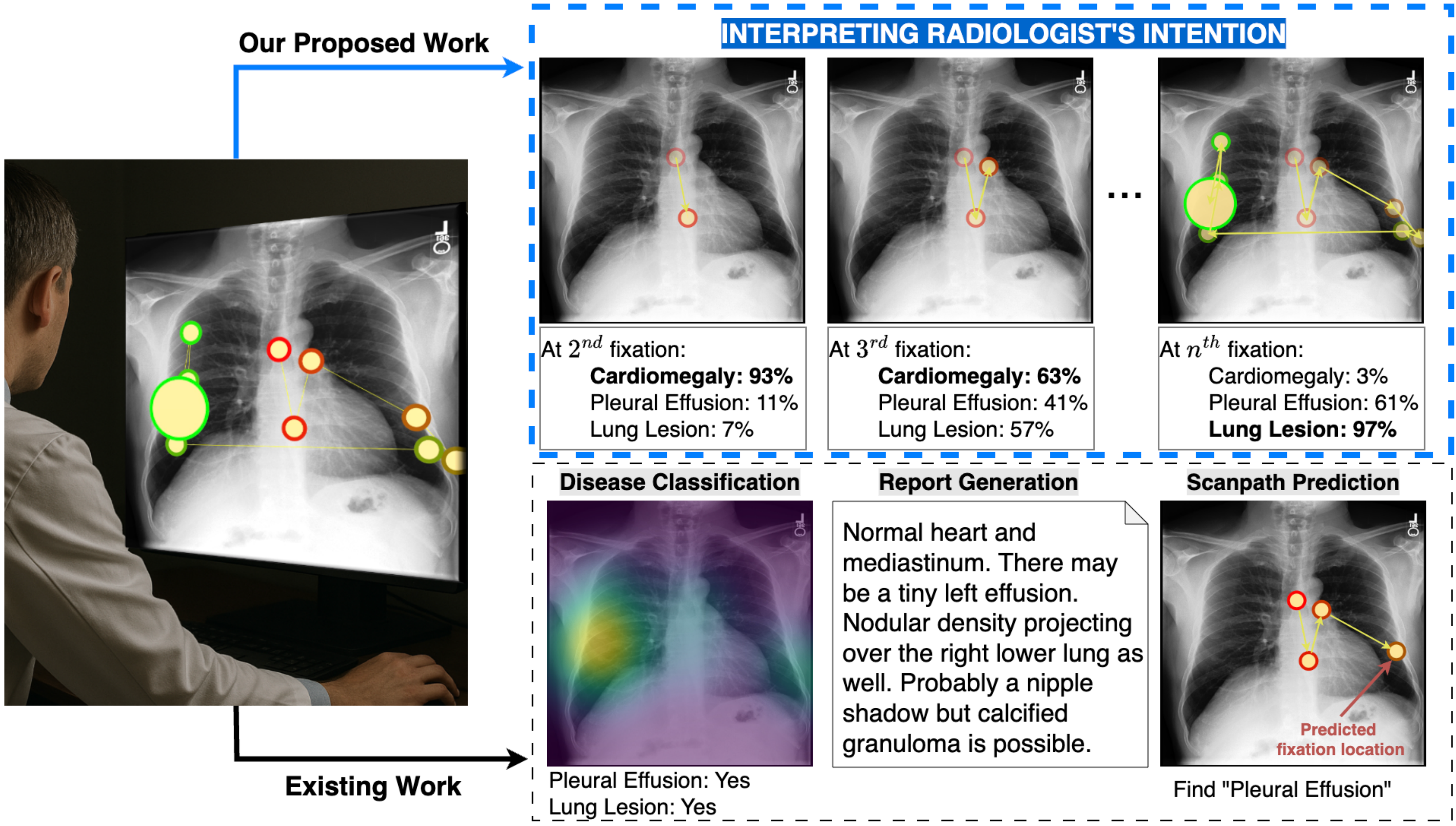}
    \caption{From existing gaze datasets (e.g., EGD and REFLACX) that are illustrated in the left image, existing research on gaze-assisted medical AI primarily focus on developing systems to assist radiologists by performing tasks such as disease classification, report generation, or mimicking visual search patterns through scanpath prediction tasks. Our work extends apart from these tasks and focuses on understanding radiologists by interpreting the radiologist's intention behind each of their captured gaze points. }
    \label{fig:teaser}
\end{teaserfigure}

\maketitle

\section{Introduction}

Radiologists rely on eye movements to navigate and interpret medical images, leading to a natural question: "what are they trying to find at this moment?" Most current works attempt to mimic radiologists to create digital twins rather than using data from them to understand them, which could be crucial in interactive systems. To the best of our knowledge, no existing work deciphers the underlying intentions behind each fixation in medical imaging analysis.

As shown in \cref{tab:comparison1}, existing Aritificial Intelligence approaches (AI) fall short in intention prediction. I-AI~\cite{pham2024ai}, despite designing models to mimic radiologist decision-making processes, only focuses on predicting heatmaps (spatial information) without explaining the purpose of each fixation. Similarly, EyeXNet~\cite{hsieh2024eyexnet}, while using fixation data, takes full heatmaps as input without considering temporal characteristics, and is used for localization rather than explaining fixation meaning. 
The work most similar to ours is ChestSearch~\cite{pham2024gazesearch}, which predicts how a radiologist would view an chest X-ray (CXR) image given a specific finding to search for. However, it is limited to a single finding, whereas in reality, radiologists look for multiple findings, and each gaze point does not always serve only one class, leading to complex relationships. Furthermore, the GazeSearch data from ChestSearch significantly reduce fixation points from the original data, which would be difficult to implement in real scenarios for an interactive system. In contrast, our RadGazeIntent model directly takes the fixation sequence and chest X-ray image to predict which finding each fixation is used to identify, along with confidence scores.

Because we strive to understand radiologists' intentions, we designe our model based on the premise that eye movements~\cite{neves2024shedding} provide a window into cognitive processes. We conceptualize intention through three complementary perspectives as illustrated in \cref{fig:teaser}: First, intention may manifest as a systematic sequential search where radiologists follow a mental checklist, targeting specific findings sequentially and perform medical visual search tasks sequentially~\cite{pham2024gazesearch}. Second, intention could reflect uncertainty-driven exploration, where radiologists respond opportunistically to visual cues without predetermined targets, in practice, this means that for a predefined set of findings, radiologists identify and report whatever they happen to observe~\cite{chen2022cocofreeview1}. Third, intention might follow a hybrid pattern where radiologists initially conduct a brief reference scan of the entire image before focusing intently on a single pathology, effectively combining the systematic and opportunistic approaches~\cite{shi:2020:air}. In this paper, we cannot completely rule out any of these possibilities, so we process the original data and evaluate all methods across multiple definitions to provide the most comprehensive perspective.
To support these different interpretations of radiologist behavior, we introduce three corresponding datasets: RadSeq (Systematic Sequential Search), RadExplore (Uncertainty-driven Exploration), and RadHybrid (Hybrid Pattern). Each dataset represents a different conceptualization of how radiologists allocate their visual attention during diagnosis.

To model the intention behind each fixation, we use a transformer-based architecture called RadGazeIntent to model gaze sequences with three major characteristics: incorporating both peripheral and foveal information to mimic the visual information humans perceive~\cite{zhibo:2020:cocosearch}, ensuring fixation information adheres to causality (earlier fixations cannot access information from later ones), and recognizing that fixations are not independent but complementary to each other, meaning adjacent fixations can be combined into more complex features. This approach allows our model to learn patterns from "fine-grained and noisy" fixation data and transform them into "coarse and abstract representations that cluster only the most relevant information." Our RadGazeIntent model employs a transformer architecture with a pooling mechanism to achieve this transformation.

Our main contributions include: 
\begin{itemize}
    \item \textbf{Benchmark:} Three new benchmark datasets, RadSeq, RadExplore, and RadHybrid, representing different conceptualizations of radiologist's intention.
    \item \textbf{RadGazeIntent:} A novel framework, RadGazeIntent, for classifying radiologists' fixations according to their underlying intentions, bridging the gap between visual search patterns and diagnostic reasoning.
    \item \textbf{Evaluation:} A comprehensive evaluation across multiple settings for predicting radiologist intentions, demonstrating our model's ability to generalize across varied intention definitions and consistently outperform baseline approaches.
\end{itemize}

\begin{table*}[t]
\centering
\caption{Comparison of previous works on eye tracking assistance methods. Most existing works using eye tracking datasets primarily utilize information in heatmap form (spatial modeling) to solve problems within their corresponding settings. Recently, ChestSearch~\cite{pham2024gazesearch} has focused on the visual search problem and proposed models to incorporate temporal information. It is also notable that a temporal classifier proposed by Karargyris et al.~\cite{karargyris2021creation} can model temporal aspects by using RNNs on heatmaps of gaze sequences. However, despite having both temporal and spatial modeling capabilities, no existing work has tackled the problem of understanding the intention behind each gaze point.}
\label{tab:comparison1}
\begin{tabular}{l|cccc}
\toprule
\multirow{1}{*}{\textbf{Methods}} & \textbf{Temporal Modeling} & \textbf{Spatial Modeling} & \textbf{Intention Interpretation} & \multirow{1}{*}{\textbf{Tasks}} \\ \toprule
 % & \textbf{Modeling} & \textbf{Modeling} & \textbf{Intention} &  \\ \hline
I-AI~\cite{pham2024ai} & \textit{\ding{55}} & \textit{\checkmark} & \textit{\ding{55}} & \textit{Disease Classification} \\
GazeRadar \cite{bhattacharya2022gazeradar} & \textit{\ding{55}} & \textit{\checkmark} & \textit{\ding{55}} & \textit{Disease Localization} \\
EGGCA-Net \cite{peng2024eye} & \textit{\ding{55}} & \textit{\checkmark} & \textit{\ding{55}} & \textit{Report Generation} \\
EyeXNet \cite{hsieh2024eyexnet} & \textit{\ding{55}} & \textit{\checkmark} & \textit{\ding{55}} & \textit{Disease Localization} \\
Karargyris et al. \cite{karargyris2021creation} & \textit{\checkmark} & \textit{\checkmark} & \textit{\ding{55}} & \textit{Disease Classification} \\
ChestSearch \cite{pham2024gazesearch} & \textit{\checkmark} & \textit{\checkmark} & \textit{\ding{55}} & \textit{Scanpath Prediction} \\ \midrule
\textbf{RadGazeIntent (Ours)} & \textit{\checkmark} & \textit{\checkmark} & \textit{\checkmark} & \textit{Intention Interpretation } \\ \bottomrule
\end{tabular}
\end{table*}
\section{Related Work}

\subsection{Gaze-Assisted Medical AI}
General gaze prediction models have been on a rise from both static saliency prediction~\cite{matthias:2016:deepgaze,marcella:2018:sam,xun:2015:salicon,camilo:2020:umsi,souradeep:2022:agdf,sen:2020:eml,shi:2023:personalsaliency,bahar:2023:tempsal} and scanpath prediction~\cite{zhibo:2023:hat,wanjie:2019:iorroi,zhenzhong:2018:iorroi-lstm,matthias:2022:deepgaze,xianyu:2021:vqa,zhibo:2020:cocosearch,zhibo:2022:targetabsent,mengyu:2023:scanpath,sounak:2023:gazeformer,xianyu:2024:individualscanpath,peizhao:2023:uniar}. For example, DeepGaze III~\cite{matthias:2022:deepgaze} and Gazeformer~\cite{sounak:2023:gazeformer} use deep learning to predict scanpaths in free-viewing tasks. Chen et al.~\cite{chen2024isp} advance this with individualized scanpath prediction. However, these methods lack adaptation to medical contexts.

Recent approaches integrate medical gaze data into AI frameworks. Karargyris et al.~\cite{karargyris2021creation} introduce a chest X-ray dataset with eye-tracking, but did not align gaze patterns with diagnostic labels. I-AI~\cite{pham2024ai} is a system decoding radiologists' focus, primarily mimicking expert attention by predicting gaze heatmap. Pham et al.~\cite{pham2024gazesearch} then introduce GazeSearch for radiology findings search, focusing on scanpath prediction rather than intention.

A key limitation across these efforts is the absence of explicit alignment between gaze sequences and diagnostic intention. They do not interpret \textit{why} a radiologist fixates on a region, whether for systematic search, uncertainty, or hybrid strategies as we do in our framework. Moreover, Neves et al.~\cite{neves2024shedding} provide a review of gaze-driven interpretability in radiology, also confirming this gap and calling for AI models that decode expert intent, a challenge that our work directly addresses.

\subsection{Multi-Label Findings Classification}
Multi-label classification of radiological findings has seen significant progress with deep learning~\cite{rajpurkar2017chexnet,huang2017densenet,yao2018weakly,taslimi2022swinchex,liu2020semi,liu2021self,gazda2021self,azizi2021big,liu2019align,yan2018weakly,li2018thoracic,van2023probabilistic}, particularly for chest X-rays. CheXNet~\cite{rajpurkar2017chexnet} demonstrated radiologist-level pneumonia detection using a convolutional neural network (CNN), while Irvin et al.~\cite{irvin2019chexpert} introduced CheXpert, a large dataset with uncertainty labels for multi-label classification. Wu et al.~\cite{wu2021chest} further advanced this with the Chest ImaGenome dataset, incorporating clinical reasoning via anatomical annotations. Transformer-based models have also been applied to this task. Taslimi et al.~\cite{taslimi2022swinchex} used Swin Transformers for multi-label chest X-ray classification, achieving robust performance across findings. Wang et al.~\cite{wang2022mgca} proposed a multi-granularity cross-modal alignment framework, integrating text and image features for generalized representation learning.
These works excel in identifying multiple pathologies but rely solely on image data, ignoring gaze-informed supervision that could reveal diagnostic priorities. 

\textit{Unlike these approaches, our model integrates gaze sequences with multi-label classification, focusing on predicting not \textbf{what} findings are present but \textbf{which} finding a radiologist is examining at a given moment.} 

\section{Methodology}
\subsection{Problem Formulation}
We formulate the task of fixation-based intention interpretation as a sequence labeling problem. Given a series of $T$ eye fixations $\mathcal{F} = \{f_1, f_2, \ldots, f_T\}$ where each fixation $f_i = (x_i, y_i, d_i)$ consists of spatial coordinates $(x_i, y_i)$ and duration $d_i$. Each fixation may be associated with one of $K$ possible intentions, reflecting the observation that multiple consecutive fixations typically correspond to a single cognitive process. We denote the ground truth intention label for each fixation as $L = \{l_1, l_2, \ldots, l_T\}$ where $l_i  \in \{0,1\}^K$. Our goal is to identify the underlying user intentions that generated these fixations.

\subsection{Architecture: RadGazeIntent}
\begin{figure*}[t]
\centering
% https://drive.google.com/file/d/1y1UBo5-JoClu4DenmdqCPaDbe-IhyR37/view?usp=sharing
\includegraphics[width=\linewidth]{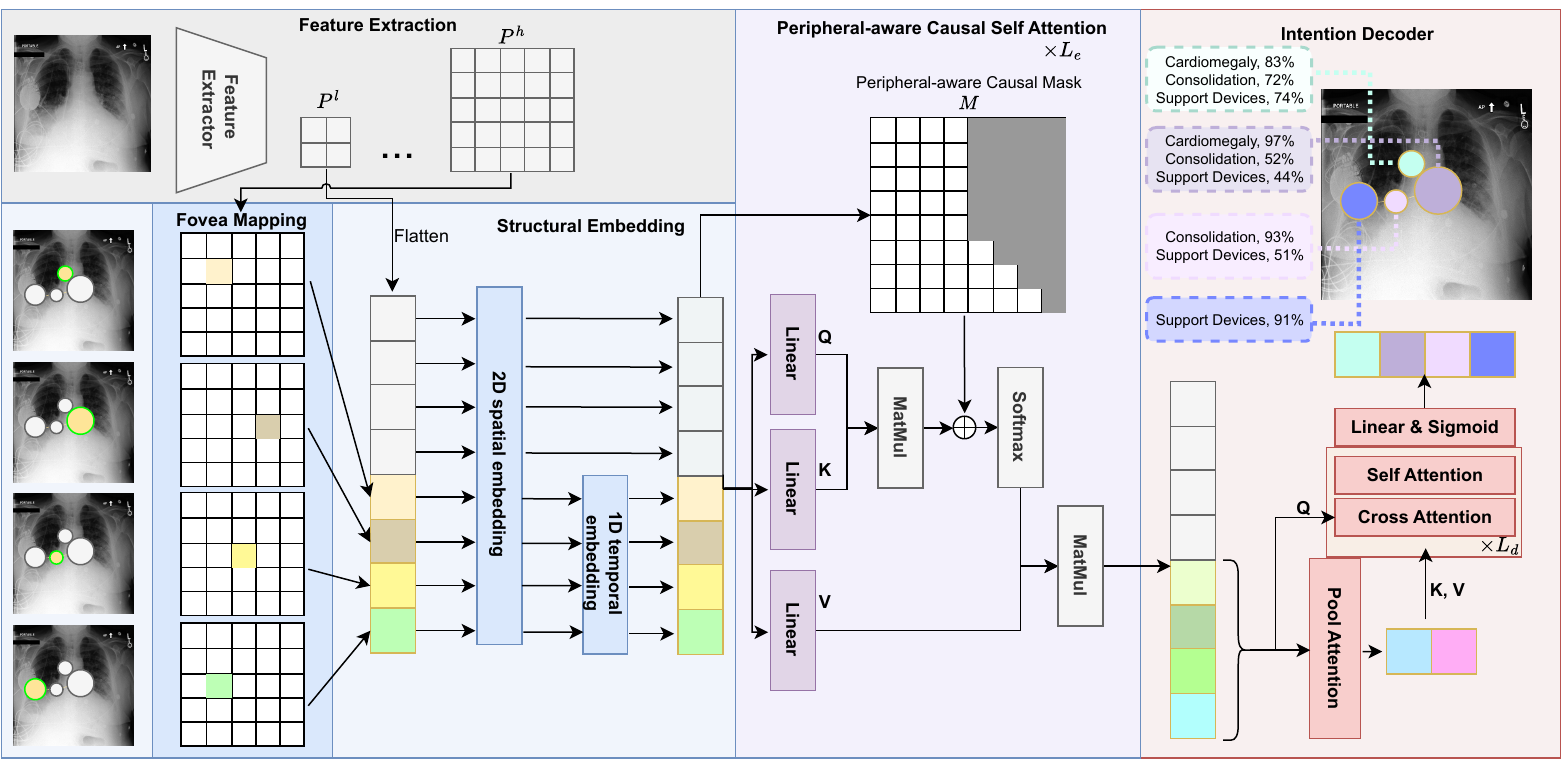}
\caption{Overall framework of RadGazeIntent. RadGazeIntent aims to analyze  medical image fixation patterns to determine diagnostic intentions. The Feature Extraction module processes a input CXR to create two distinct feature maps, peripheral ($P^l$) for general context and fovea ($P^h$) for detailed focus areas. The Structural Embedding module transforms fixation coordinates into feature representations using the fovea feature map ($P^h$). This module incorporates both spatial (2D) and temporal (1D) structural information to maintain the relationship between fixation points. Then the Peripheral-aware Causal Self Attention is a specialized attention mechanism that enables the model to learn features for each fixation in context. It references peripheral-level image information while preserving causality, i.e. ensuring that earlier fixations cannot access information from later ones, with the help of Peripheral-aware Causal Mask $M$ (see \cref{eq:mask}). Our implementation stacks $L_e$ blocks of this attention mechanism. The Intention Decoder uses a Pool Attention module to condense the token representation. These condensed tokens are processed through $L_d$ blocks of cross-attention and self-attention layers, followed by linear and sigmoid layers. This produces confidence scores for specific diagnostic findings, such as "Cardiomegaly?, "Consolidation" and "Support Devices," as shown in the top right of this figure.}
\label{fig}
\end{figure*}

The proposed framework processes a CXR through a sequential pipeline as shown in \cref{fig}: Starting with CXR images and fixation data, the Feature Extraction module generates peripheral and foveal feature maps. These features are passed to the Structural Embedding module, which encodes spatial and temporal characteristics of the fixation sequence. The embedded features are then processed by the Peripheral-aware Causal Self-Attention mechanism to integrate contextual dependencies. Finally, the Intention Decoder classifies the diagnostic purpose of each fixation and outputs corresponding confidence scores for relevant radiological findings.

\noindent
\textbf{Feature Extraction.}
The goal of this step is to create a feature pyramid to represent both peripheral visual information and fovea visual information of an image. For an input image $I \in \mathbb{R}^{H \times W}$, we use a Feature Pyramid Network (FPN)~\cite{lin2017fpn} with ResNet~\cite{he2015resnet} backbone to obtain pyramid features $\{P^1, P^2, P^3, P^4\}$ with varying resolutions. We select $P^l=P^1 \in \mathbb{R}^{C \times H/32 \times W/32}$ with the lowest resolution to represent peripheral visual information and $P^h=P^4 \in \mathbb{R}^{C \times H/4 \times W/4}$ with the highest resolution to represent fovea visual information.

\noindent
\textbf{Structural Embedding.}
This component embeds the fixation sequence into meaningful features that represent: (1) feature relevance to image content, (2) 2D spatial properties, (3) temporal sequence information. First, we perform Fovea Mapping by extracting features from the corresponding spatial locations $(x_i, y_i)$ in the fovea feature map $P^h$, mapping a point $(x,y)$ to the feature at block $(x/4,y/4)$ based on the scale of $P^h$. In total, we obtain $E_f = \{e_1, e_2, \dots, e_T\} \in \mathbb{R}^{C \times T}$ corresponding to $T$ input fixations, wehre $e_i \in \mathbb{R}^C$ is the embedded feature of the $i^{th}$ fixation.

Then, we apply 2D Spatial Embedding~\cite{cheng2022masked} to embed 2D spatial information into the feature representation using sinusoidal functions, followed by 1D Temporal Embedding~\cite{vaswani2017attention} which incorporates sequence order information by encoding the sequential position $t$ of each fixation point. After embedding, we obtain structural embedded fixation feature $\bar{E}_f$. Simultaneously, we flatten $P^l$ into multiple tokens $\bar{P}^l \in \mathbb{R}^{C \times (HW/1024)}$ and also apply 2D spatial embedding to these tokens. We then concatenate the embeddings into a single vector $E_s = [\bar{P}^l, \bar{E}_f] \in \mathbb{R}^{C \times (HW/1024 + T)}$ and pass it to the next step.

\noindent
\textbf{Peripheral-aware Causal Self Attention.}
This specialized self-attention mechanism incorporates peripheral information while maintaining causality in the sequence processing. It uses a Peripheral-aware Causal Mask that allows each fixation to access all preceding fixations and peripheral information but prevents access to future fixations. As we have a total of $HW/1024 + T$ tokens, we need a mask $M$ with the size of $(HW/1024 + T) \times (HW/1024 + T)$. We create Peripheral-aware Causal Mask as:
\begin{equation}
\text{M}_{ij} =
\begin{cases}
0 & \text{if } i > j \text{ or } j \in [1,HW/1024] \\
-\infty & \text{otherwise}
\end{cases}
\label{eq:mask}
\end{equation}
where $i,j \in [1, HW/1024 + T]$ corresponding to row and column indexes. 
\begin{equation}
\text{Attention}(\textbf{Q}, \textbf{K}, \textbf{V}) = \text{softmax}\left(\frac{\textbf{Q}\textbf{K}^T + \text{M}}{\sqrt{d_k}}\right)V
\end{equation}
where $\textbf{Q}$  is the query, $\textbf{K}$ is the key, $\textbf{V}$ is the value, created by passing $E_s$ through separate Linear layers~\cite{vaswani2017attention}, $M$ is the Peripheral-aware Causal Mask, and $d_k$ is the hidden dimension of $\textbf{K}$.

This self-attention block transforms our encoded features into a deep latent space representing both image and fixation information simultaneously. Multiple ($L_e$) layers of this attention mechanism are applied to capture complex relationships. After this step, we obtain the contextualized feature $E_s' = [\bar{P}^{l'}, E_f']$, where  $\bar{P}^{l'}$ is a global information and $E_f'$ is the fixation feature. In this module, $\bar{P}^{l'}$ aims to enrich context information to $E_f'$, so we split and only use $E_f'$ for the next step.

\noindent
\textbf{Intention Decoder.}
The role of the Intention Decoder is to explicitly filter out noise and capture more complex patterns that represent intentions and then decode that features into the intention behind each fixation.
Intuitively, intentions typically span multiple fixations. Thus, we use Pool Attention to compress the feature sequence $E_f'$, reducing from $T$ tokens to fewer tokens, $E_f^*$ . Next, Self-Attention and Cross-Attention layers allow each feature token from $E_f'$ to query and select the most appropriate latent features representing underlying intentions from $E_f^*$. This process is repeated for $L_d$ layers. Finally, a Linear layer transforms the decoder output into the intention space, and a Sigmoid layer normalizes values between 0 and 1, providing confidence scores for specific findings such as "Cardiomegaly," "Consolidation," and "Support Devices" as shown in the \cref{fig}.

\noindent
\textbf{Objective Function.}
The proposed problem of fixation interpretation can be formulated as a multi-label classification task, so we use binary cross-entropy as our loss function:
\begin{equation}
\mathcal{L} = -\frac{1}{TK}\sum_{i=1}^{T}\sum_{k=1}^{K} \left[ l_{ik} \log(\hat{l}_{ik}) + (1-l_{ik}) \log(1-\hat{l}_{ik}) \right]
\end{equation}
where $l_i$ is the ground truth label, $\hat{l}_i$ is the predicted probability. 

\section{Experiments}
\subsection{Datasets}
\label{sec:dataset}
\begin{figure}[t]
    \centering
    % https://drive.google.com/file/d/12axLSzvX_mTt0jkWhrzDpWBkXaiLfJ8m/view?usp=sharing
    \includegraphics[width=\linewidth]{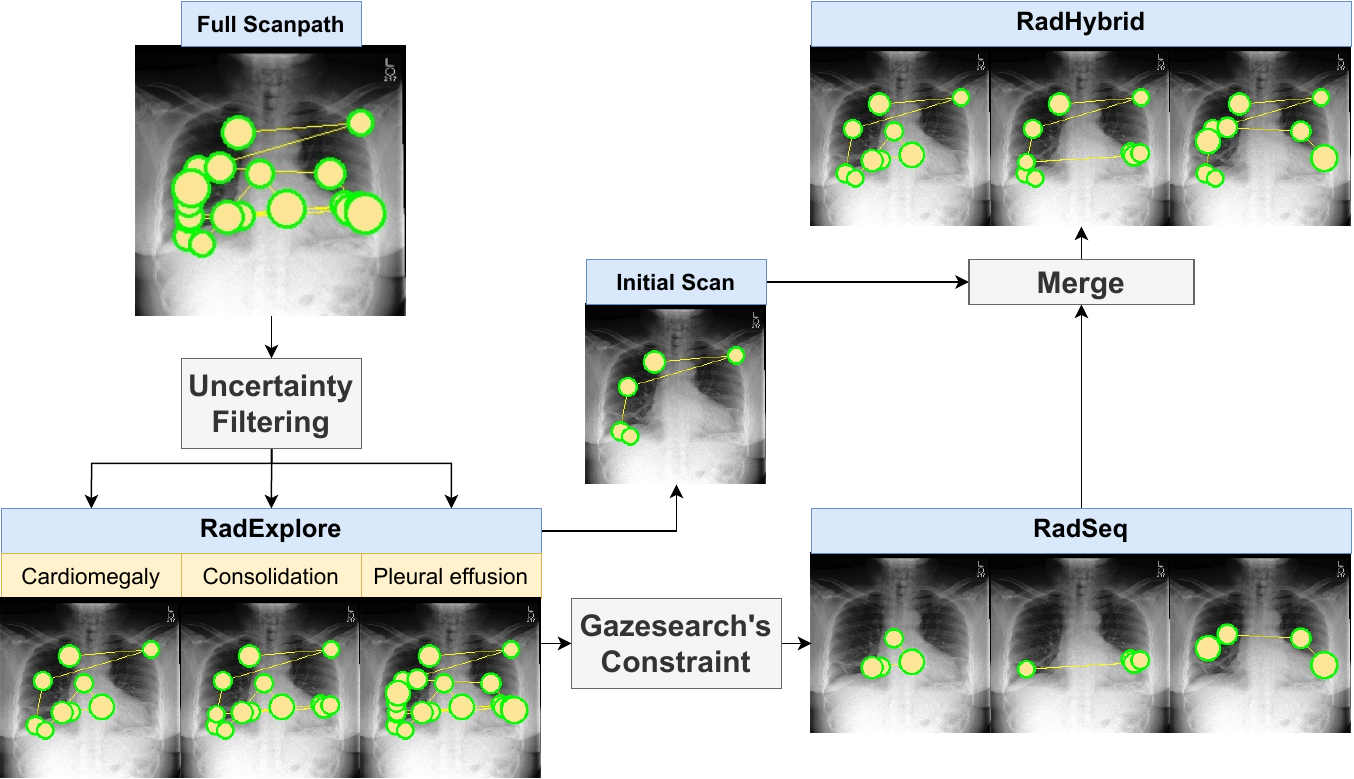}
    \caption{Illustration of our dataset creation process, which transforms the original eye-tracking data into three distinct experimental settings: RadSeq (Systematic Sequential Search), RadExplore (Uncertainty-driven Exploration), and RadHybrid (Hybrid Pattern). Beginning with a full scanpath showing radiologist fixation points (green and yellow) on a chest X-ray, we apply \textit{uncertainty filtering} to isolate fixations that fall outside annotated findings, forming RadExplore, which models exploratory behavior under diagnostic uncertainty. In this example, we extract fixations for three findings: Cardiomegaly, Consolidation, and Pleural Effusion. Next, we implement Gazesearch’s constraints~\cite{pham2024gazesearch} to convert RadExplore into pathology-focused fixations, creating RadSeq, which simulates systematic and targeted search. To construct RadHybrid, we merge the extracted \textit{initial scan} from the first few seconds (see \cref{sec:dataset}) with RadSeq to capture a behavior pattern that begins with broad scanning and transitions into focused searching.}
    \label{fig:systematic_search}
\end{figure}

\begin{figure}[t]
\centering
\begin{subfigure}[b]{0.49\linewidth}
    \centering    \includegraphics[width=\linewidth]{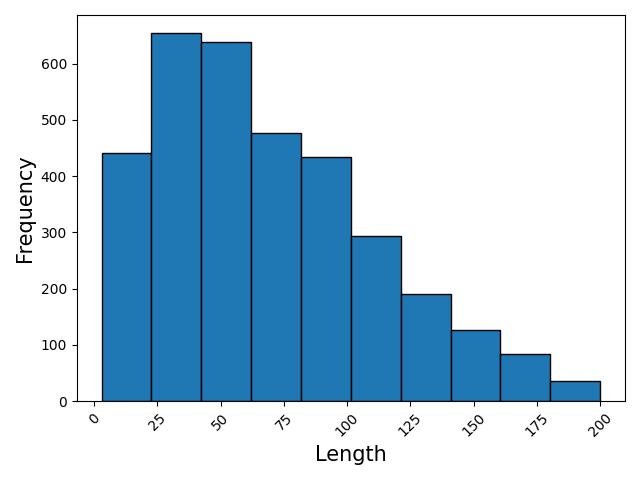}
    \caption{Fixation Length Distribution.}
    \label{fig:setting4}
\end{subfigure}
\hfill
\begin{subfigure}[b]{0.49\linewidth}
    \centering
    \includegraphics[width=\linewidth]{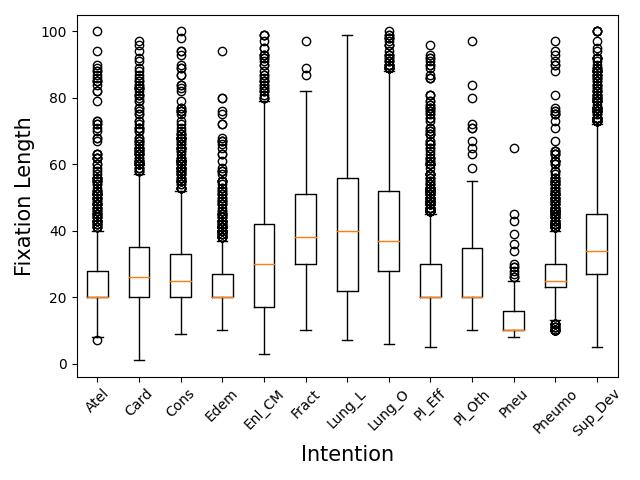}
    \caption{RadSeq}
    \label{fig:setting1}
\end{subfigure}
\vskip\baselineskip
\begin{subfigure}[b]{0.49\linewidth}
    \centering
    \includegraphics[width=\linewidth]{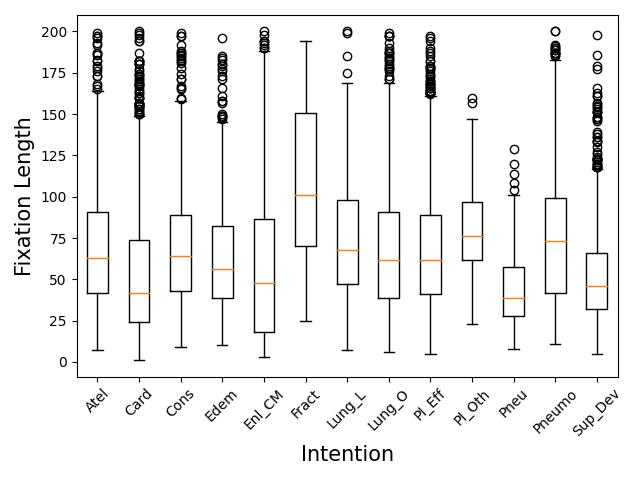}
    \caption{RadExplore}
    \label{fig:setting2}
\end{subfigure}
\hfill
\begin{subfigure}[b]{0.49\linewidth}
    \centering
    \includegraphics[width=\linewidth]{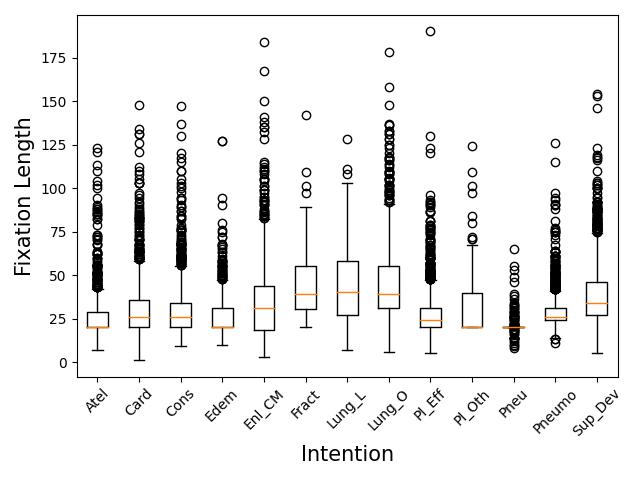}
    \caption{RadHybrid}
    \label{fig:setting3}
\end{subfigure}
\caption{Statistical analysis of eye fixation patterns across different datasets. (a) Histogram showing the overall distribution of fixation lengths, with most fixations concentrated between 25-75 points. (b-d) Box plots displaying fixation length distributions across 13 radiological findings for three different datasets: RadSeq, RadExplore, and RadHybrid. Each box plot represents the median, interquartile range, and outliers of fixation lengths for specific medical findings. The abbreviated labels correspond to: Atel (Atelectasis), Card (Cardiomegaly), Cons (Consolidation), Edem (Edema), Enl\_CM (Enlarged Cardiomediastinum), Fract (Fracture), Lung\_L (Lung Lesion), Lung\_O (Lung Opacity), Pl\_Eff (Pleural Effusion), Pl\_Oth (Pleural Other), Pneu (Pneumonia), Pneumo (Pneumothorax), and Sup\_Dev (Support Devices).}
\label{fig:all_settings}
\end{figure}

To investigate the intention behind each radiologist's gaze, we derive three intention-labeled datasets by post-processing two publicly available gaze datasets, EGD~\cite{karargyris2021creation} and REFLACX~\cite{bigolin2022reflacx}, under distinct behavioral assumptions that reflect plausible visual search strategies~\cite{chen2022cocofreeview1,pham2024gazesearch,bertram2013effect}. \cref{fig:systematic_search} illustrates the overview of our data processing steps. The three newly introduced intention-labeled datasets are as follows:

\noindent
\textbf{RadExplore: Uncertainty-Driven Exploration.} This dataset considers intention as opportunistic visual search~\cite{chen2022cocofreeview1}, assuming radiologists do not follow a fixed order and may consider all report findings simultaneously. 
This reflects maximal ambiguity: all fixations are potentially relevant to any finding, leaving intention disambiguation to later modeling stages. Formally, let $S = \{s_1, s_2, ..., s_{|S|}\}$ be the sequence of sentences in the transcript, where each sentence has an end time $s_j^e$, and $\{\tau_i\}_{i=1}^T$ is the set of captured timestampt for $\{f_i\}_{i=1}^T$ fixations. 
We use CheXbert~\cite{smit2020chexbert} to find the corresponding class for all sentences and produce $C = \{c_1, c_2, \dots, c_{|S|}\}$. Then we compute the ground truth labels as:
\begin{equation}
\mathbf{l}_{ik} = \begin{cases}
    1 &  \text{if there exists } j \in [1, |S|] \text{ such that } \tau_i \leq s_j^e \text{ and } k = c_j\\ 
    0 & \text{otherwise}
\end{cases}
\end{equation}
where $i \in [1, T]$ indexes fixations and $k \in [1, K]$ indexes intention labels. We refer to this step as Uncertainty Filtering.

\noindent
\textbf{RadSeq: Systematic Sequential Search.} This dataset assumes radiologists follow a sequential checklist of findings~\cite{pham2024gazesearch} and solely focus on searching clues for a particular finding at a time. Gazesearch's constraints~\cite{pham2024gazesearch} comprise of two procedures: radius-based filtering and time-spent constraining.
Using these constraints on RadExplore, we obtain the beginning time $\text{beg}_k$ and end time $\text{end}_k$ for all $K$ intentions in the report $S$.
Then we compute the ground truth labels as:
\begin{equation}
\mathbf{l}_{ik} = \begin{cases}
    1 & \text{if } \tau_i \in [\text{beg}_k,\text{end}_k]\\ 
    0 & \text{otherwise}
\end{cases}
\end{equation}
Unlike the original paper~\cite{pham2024gazesearch}, we set the radius to zero in the radius-based filtering procedure to avoid discarding any fixation points, thereby preventing a reduction in the temporal and spatial information of the fixations.

\noindent
\textbf{RadHybrid: Hybrid Pattern.} According to \cite{bertram2013effect}, a radiologist's intention leads to a two-phase search process where radiologists begin with a broad overview and later narrow focus to specific pathologies. To get the "broad overview" behavior, we extract the scanning fixations within the first $\tau^*$ seconds. Often, $\tau^*$ is 1 second according to \cite{shi:2020:air}.
Finally, we merge initial scanning fixations with RadSeq:
\begin{equation}
l_{ik} = 
\begin{cases}
1 & \text{if } \tau_i \leq \tau^* \text{ or } \tau_i \in [\text{beg}_k,\text{end}_k]\\
0 & \text{otherwise}
\end{cases}
\end{equation}
where $\tau^*=1$ is the initial scanning time. 

Each of the three datasets, RadSeq, RadExplore, and RadHybrid, is derived from the same source of eye tracking data, i.e., 1,079 samples from EGD and 2,483 samples from REFLACX. \cref{fig:all_settings} illustrates the patterns in radiologists' eye fixation behavior across different radiological findings. The histogram shows most fixations are relatively brief (25-75 points), though with some longer outliers. Across all three datasets (RadSeq, RadExplore, and RadHybrid), certain conditions consistently demand longer fixation lengths, particularly Lung Lesions and Fractures, which stand out with higher median values and wider interquartile ranges, suggesting these conditions may require longer visual attention during diagnosis.

\subsection{Compared Baselines}
We compare our approach against several baselines representing different architectural paradigms:
We compare our model against a range of existing approaches that capture different modeling assumptions for intention prediction from radiologist gaze data.

\noindent\textbf{MLP.} This baseline uses a multilayer perceptron that processes each fixation independently. Each input fixation is first mapped with fovea feature like in our framework. Then they are passed through three fully connected layers (512, 256, $K$ units) with ReLU activations, where $K$ denotes the number of intention classes. Finally the latent features are passed through a sigmoid layer for multi-label classification. This model lacks any temporal modeling or spatial aggregation over time. 

\noindent\textbf{LSTM.} A sequential model that encodes temporal dynamics in gaze behavior using a unidirectional LSTM. Similar to MLP, we also use mapped fovea features to represent the fixation token features. Then we use the LSTM decoder on the fixation token features.
The model has 256 hidden units with a dropout rate of 0.2 between layers. The final hidden state is projected to the intention space via a fully connected layer. Finally the projected features are passed through a sigmoid layer for multi-label classification.

\noindent\textbf{Karargyris et al.~\cite{karargyris2021creation}.} This model first transforms fixation sequences into spatial heatmaps using Gaussian kernels. The input CXR and these heatmaps are passed through a ResNet-18 CNN encoder, followed by a bidirectional LSTM (256 hidden units), a temporal convolutional layer, and a final classification head. Unlike the original implementation in \cite{karargyris2021creation}, we modify the final classification head: instead of predicting three classes, we apply the classification head separately for each token to get the multi-label prediction.

\noindent\textbf{ChestSearch~\cite{pham2024gazesearch}.} This baseline is originally designed for the GazeSearch dataset. We change the final decoder heads of ChestSearch from decoding heatmaps to a classifier for predicting intention.

All deep learning models are trained with the Adam optimizer, initial learning rate of 1e-4 with cosine annealing schedule, and for the same number of epochs (100) with early stopping based on validation performance to ensure fair comparison.

\subsection{Implementation Details}
We use a Pyramid Feature Network with ResNet-50 backbone~\cite{he2015resnet} as the Feature Extractor, initialized from a checkpoint pre-trained using MGCA~\cite{wang2022mgca} for 50 epochs with a batch size of 144. This Feature Extractor is frozen when we train the full pipeline.
We stack $L_e = 4$ Peripheral-aware Causal Self Attention layers with hidden size $D = 384$ and $H = 8$ attention heads.

To reduce sequence length and retain complex features, we apply a pooling attention layer~\cite{fan2021multiscale} with stride of 2 and kernel size of 5 tokens. The Intention Decoder contains $L_d = 6$ blocks of self-attention and cross-attention. We train the entire model for 4,000 iterations using the AdamW optimizer~\cite{loshchilov2017adamw}, with a learning rate of $1 \times 10^{-5}$ and a batch size of 32. All experiments are conducted on a single NVIDIA A6000 GPU with 48GB of RAM.

We then evaluate intention predictors using a set of classification metrics, i.e., Accuracy (ACC), F1-score (F1), Precision (P), and Recall (R), for every pair of fixation-intention. We run 5-fold cross validation and report 95\% confidence interval in \cref{sec:quant}.

\subsection{Quantitative Results}
\label{sec:quant}
\begin{table*}[t]
\centering
\caption{Performance comparison across two datasets (EGD and REFLACX) with 95\% confidence intervals ($\pm$). Metrics include Accuracy (ACC), F1-score (F1), Precision (P), and Recall (R).}
\label{tab:comparison}
\resizebox{\linewidth}{!}{%
\begin{tabular}{l|l|cccc|cccc}
\toprule
\multirow{2}{*}{\backslashbox{\textbf{Datasets}}{\textbf{Data Sources}}} & \multirow{2}{*}{\textbf{Model}} & \multicolumn{4}{c|}{\textbf{EGD}} & \multicolumn{4}{c}{\textbf{REFLACX}} \\
&  & \textbf{ACC (\%)} & \textbf{F1 (\%)} & \textbf{P (\%)} & \textbf{R (\%)} & \textbf{ACC (\%)} & \textbf{F1 (\%)} & \textbf{P (\%)} & \textbf{R (\%)} \\
\midrule

\multirow{6}{*}{\textbf{RadSeq}} 
& MLP           & 73.87 ($\pm$1.3) & 49.14 ($\pm$1.6) & 62.16 ($\pm$1.5) & 51.40 ($\pm$1.8) & 82.55 ($\pm$1.2) & 52.33 ($\pm$1.4) & 58.90 ($\pm$1.6) & 54.76 ($\pm$1.3) \\
& LSTM          & 81.23 ($\pm$1.5) & 56.77 ($\pm$1.4) & 59.12 ($\pm$1.3) & 54.89 ($\pm$1.7) & 79.98 ($\pm$1.7) & 55.21 ($\pm$1.5) & 60.01 ($\pm$1.3) & 53.43 ($\pm$1.4) \\
& Karargyris et al.    & 84.02 ($\pm$1.2) & 61.88 ($\pm$1.2) & 64.77 ($\pm$1.1) & 59.45 ($\pm$1.0) & 81.22 ($\pm$1.1) & 59.34 ($\pm$1.2) & 63.15 ($\pm$1.1) & 56.70 ($\pm$1.3) \\
& ChestSearch           & 87.35 ($\pm$1.0) & 68.20 ($\pm$1.0) & 70.11 ($\pm$1.2) & 66.30 ($\pm$1.1) & 85.02 ($\pm$1.0) & 65.44 ($\pm$1.0) & 68.21 ($\pm$1.2) & 63.70 ($\pm$1.1) \\
& \textbf{Ours} & \textbf{88.85 ($\pm$0.7)} & \textbf{72.05 ($\pm$0.9)} & \textbf{74.01 ($\pm$1.0)} & \textbf{70.51 ($\pm$0.9)} & \textbf{86.92 ($\pm$0.9)} & \textbf{69.87 ($\pm$0.9)} & \textbf{72.12 ($\pm$1.0)} & \textbf{67.90 ($\pm$0.9)} \\
\midrule

\multirow{6}{*}{\textbf{RadExplore}} 
& MLP           & 72.45 ($\pm$1.4) & 50.32 ($\pm$1.5) & 59.20 ($\pm$1.6) & 52.10 ($\pm$1.3) & 81.00 ($\pm$1.3) & 51.44 ($\pm$1.5) & 56.70 ($\pm$1.3) & 50.33 ($\pm$1.4) \\
& LSTM          & 80.33 ($\pm$1.6) & 53.88 ($\pm$1.3) & 57.78 ($\pm$1.5) & 52.04 ($\pm$1.4) & 78.45 ($\pm$1.5) & 54.11 ($\pm$1.3) & 58.20 ($\pm$1.3) & 51.80 ($\pm$1.3) \\
& Karargyris et al.    & 83.12 ($\pm$1.2) & 60.45 ($\pm$1.1) & 62.99 ($\pm$1.1) & 58.77 ($\pm$1.0) & 80.76 ($\pm$1.3) & 57.44 ($\pm$1.3) & 61.02 ($\pm$1.2) & 55.21 ($\pm$1.1) \\
& ChestSearch           & 86.44 ($\pm$0.9) & 66.10 ($\pm$1.1) & 68.30 ($\pm$1.2) & 64.55 ($\pm$1.0) & 84.01 ($\pm$1.0) & 63.11 ($\pm$1.0) & 66.90 ($\pm$1.1) & 61.22 ($\pm$1.0) \\
& \textbf{Ours} & \textbf{87.95 ($\pm$0.8)} & \textbf{70.14 ($\pm$0.9)} & \textbf{72.25 ($\pm$1.0)} & \textbf{68.01 ($\pm$0.9)} & \textbf{85.40 ($\pm$0.8)} & \textbf{67.33 ($\pm$0.9)} & \textbf{70.89 ($\pm$1.0)} & \textbf{65.92 ($\pm$0.9)} \\
\midrule

\multirow{6}{*}{\textbf{RadHybrid}} 
& MLP           & 72.12 ($\pm$1.3) & 48.99 ($\pm$1.4) & 60.45 ($\pm$1.5) & 50.77 ($\pm$1.3) & 80.01 ($\pm$1.3) & 50.22 ($\pm$1.4) & 55.10 ($\pm$1.2) & 49.85 ($\pm$1.4) \\
& LSTM          & 79.80 ($\pm$1.5) & 52.77 ($\pm$1.3) & 55.89 ($\pm$1.4) & 50.43 ($\pm$1.5) & 77.33 ($\pm$1.4) & 51.10 ($\pm$1.3) & 56.01 ($\pm$1.2) & 48.88 ($\pm$1.2) \\
& Karargyris et al.    & 83.56 ($\pm$1.1) & 58.60 ($\pm$1.1) & 61.90 ($\pm$1.2) & 56.45 ($\pm$1.0) & 79.66 ($\pm$1.2) & 56.77 ($\pm$1.1) & 60.44 ($\pm$1.3) & 54.90 ($\pm$1.1) \\
& ChestSearch           & 86.22 ($\pm$0.9) & 65.77 ($\pm$1.0) & 67.91 ($\pm$1.1) & 63.80 ($\pm$1.0) & 83.45 ($\pm$0.9) & 62.70 ($\pm$1.0) & 65.11 ($\pm$1.0) & 60.88 ($\pm$1.0) \\
& \textbf{Ours} & \textbf{88.21 ($\pm$0.8)} & \textbf{71.11 ($\pm$0.8)} & \textbf{73.20 ($\pm$0.9)} & \textbf{69.88 ($\pm$0.8)} & \textbf{86.02 ($\pm$0.8)} & \textbf{68.44 ($\pm$0.9)} & \textbf{71.55 ($\pm$0.9)} & \textbf{66.78 ($\pm$0.8)} \\
\bottomrule
\end{tabular}%
}
\end{table*}

\cref{tab:comparison} presents the quantitative performance of our proposed model compared to four baseline methods (MLP, LSTM, Karargyris et al., and ChestSearch) across two eye-tracking sources (EGD and REFLACX), evaluated under three datasets representing different intention perspectives: RadExplore, RadSeq, and RadHybrid.

In RadSeq, the higher F1-scores (72.05\% for EGD and 69.87\% for REFLACX) reflect the model’s ability to accurately capture the radiologist’s systematic sequential scanning. Baseline models, particularly simpler ones like MLP and LSTM, struggle to model the temporal order of fixations, resulting in lower accuracy and recall. Thanks to our transformer-based architecture, RadGazeIntent effectively models these sequential dependencies.

As shown in \cref{fig:all_settings}, RadExplore exhibits much longer fixation sequences than RadSeq, requiring a greater number of predictions. Despite this, \textit{RadGazeIntent} continues to demonstrate robustness. For instance, its precision scores (72.25\% for EGD and 70.89\% for REFLACX) suggest that our model better infers the intent behind each fixation, even amidst uncertainty, compared to the baselines.

Finally, in RadHybrid, our model consistently outperforms all baselines across both datasets, with notable gains in recall (69.88\% for EGD and 66.78\% for REFLACX), highlighting its ability to capture both broad overview and focus scanning phases. Baseline models, except ChestSearch, lack mechanisms to disentangle these phases, leading to lower F1-scores as they conflate coarse and fine-grained fixations. Our model's improvement over ChestSearch is attributed to its pooling mechanism, which enables the separation of exploratory and focused patterns from the input fixations.

Overall, the quantitative results validate the effectiveness of our framework across all three intention definitions, establishing a strong benchmark for intention interpretation.

\begin{figure*}[t]
    \centering
    \includegraphics[width=\linewidth]{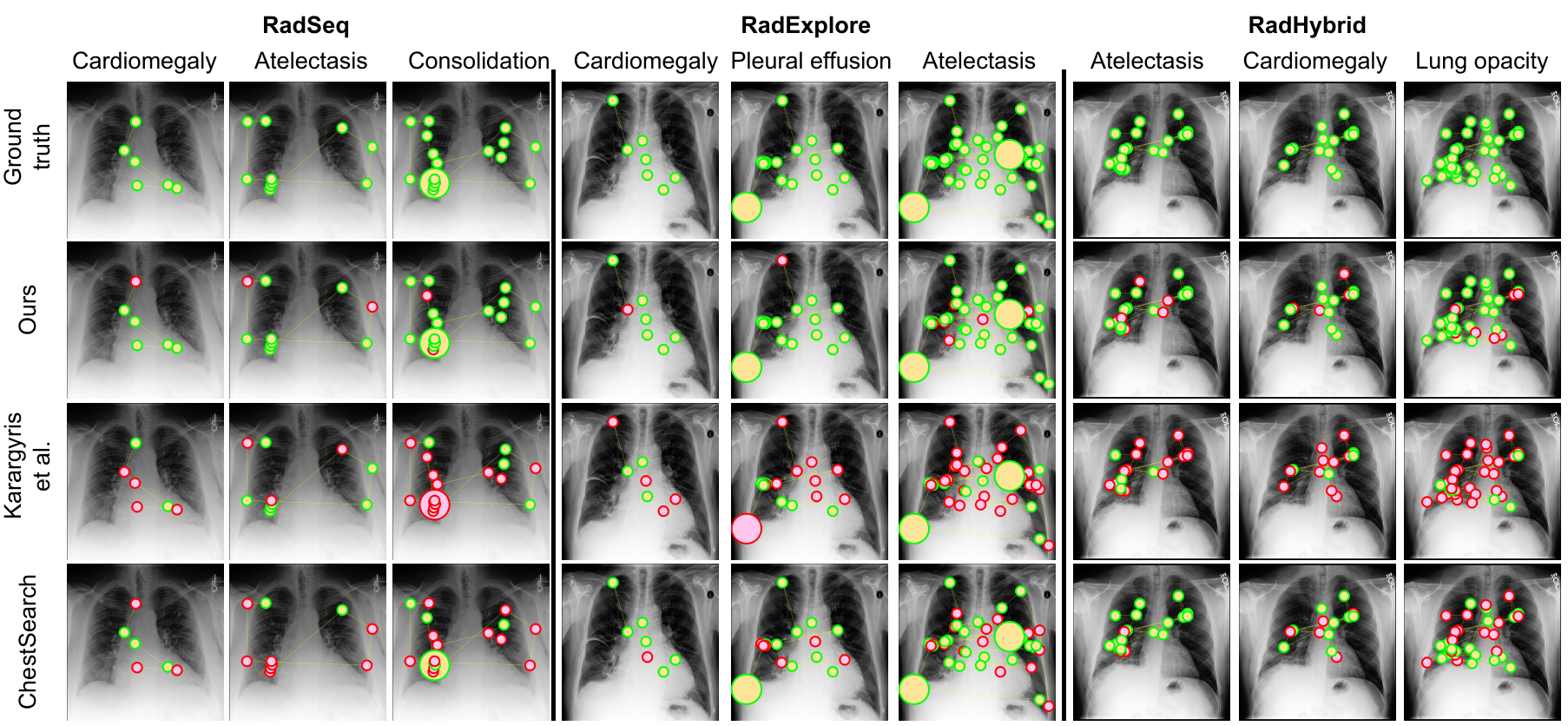}
    \caption{Qualitative comparison of radiologist intention prediction results across three datasets and various intention classes. The visualization presents chest X-rays with overlaid fixation points organized in a matrix format where columns represent different pathological findings across datasets, and rows represent different prediction methods: Ground truth, Our proposed model, Karargyris et al.'s approach, and ChestSearch. The input for the model is the full fixation sequence, and model's objecive is to predict the intention class for each point. \tikzcircle[green, fill=yellow]{4pt} represents correct prediction (i.e., confidence score of a fixation is greater than 0.5) and  \tikzcircle[red, fill=pink]{4pt} denotes incorrect prediction.} %For example, in the Cardiomegaly column, if the confidence score of a fixation is greater than 0.5, we assign "yes", otherwise "no". If it's "yes", we color it \hl{fixation?} green to represent a correct prediction; if "no", we color it red to represent an incorrect prediction.}
    \label{fig:qual1}
\end{figure*}
\subsection{Qualitative Results}

In this section, we present a qualitative analysis of our model's ability to interpret radiologists' fixation patterns and predict their underlying intentions during medical image analysis.

Figure~\ref{fig:qual1} presents a qualitative comparison of radiologist intention prediction results across three experimental settings and various intentions. Our proposed model demonstrates superior performance compared to both baseline approaches (Karargyris et al. and ChestSearch), with prediction closely resembling ground truth. This is evidenced by the prevalence of green fixation points (correct predictions) in our model's results across all settings, while the competing approaches show considerably more red indicators (incorrect predictions).
RadExplore and RadHybrid display denser fixation sequences overall, but our performance is still maintained and remains similar to the case with the fewest points, which is RadSeq. These qualitative results reinforce the quantitative advantage of our approach in accurately predicting radiologists' diagnostic intentions.
\subsection{Ablation Study}

Table~\ref{tab:ablation-components} presents an ablation study to assess the impact of removing key components from our full model, providing insights into the importance of each element in capturing radiologists' gaze intentions. The study examines the effects of removing Pool Attention, 1D Temporal Embedding, 2D Spatial Embedding, Peripheral Features, and Fovea Mapping, highlighting their individual roles in the model's ability to accurately interpret gaze patterns. 

\noindent \textbf{w/o Pool Attention.} In this setting, we remove the Pool Attention layer and use $E_f'$ directly in the Cross Attention module. This leads to a moderate drop in performance (1.29–2.81\%), highlighting its role in aggregating features that capture overall intent patterns across fixations.

\noindent \textbf{w/o 1D Temporal Embedding.} We remove the 1D Temporal Embedding block from the Structural Embedding module. Eliminating temporal encoding removes explicit modeling of fixation order, thereby discarding the temporal dynamics of visual attention. This results in a significant performance decline (4.34–4.68\%), confirming that temporal progression is vital for understanding intention.

\noindent \textbf{w/o 2D Spatial Embedding.} We remove the 2D Spatial Embedding block from the Structural Embedding module. Without spatial embedding, the model no longer receives location-aware cues about where fixations occur on the image. The performance drop (3.24–6.67\%) indicates that spatial context is also important.

\noindent \textbf{w/o Peripheral Feature.} This configuration excludes the coarse-resolution features $P^l$. As a result, the Peripheral-aware Causal Self Attention blocks perform causal self-attention solely on fixation features, without incorporating the broader contextual features from $P^l$. The performance drops (4.76–6.67\%) suggest that peripheral information, though less detailed than foveal features, contributes to understanding scene context and supports intention interpretation.

\noindent \textbf{w/o Fovea Mapping.} In this setting, we replace the Fovea Mapping with a 2D layout embedding based on coordinates~\cite{xu2020layoutlm}, commonly used in document understanding. This effectively removes the high-resolution foveal features $P^h$ and replaces them with standard 2D spatial embeddings. This leads to the most severe degradation (11.73–12.12\%), indicating that high-resolution features are crucial for identifying subtle diagnostic cues, especially during fine-grained examination phases. Their absence impairs the model’s ability to localize fixation intent around medically salient regions.

These results demonstrate that spatial and temporal cues, along with multi-scale visual processing (foveal and peripheral), are essential for effectively interpreting radiologists’ gaze intentions.

\begin{table}[h]
\centering
\caption{Ablation study on architectural components across different intention datasets. Metric: The average of F1-score (\%) on both eye tracking sources (EGD and REFLACX).}
\label{tab:ablation-components}
\resizebox{\linewidth}{!}{
\begin{tabular}{lccc}
\toprule
\textbf{Model Configuration} & \textbf{RadSeq} & \textbf{RadExplore} & \textbf{RadHybrid} \\
\midrule
Full Model & 71.01 & 68.74 & 69.78 \\ \midrule
w/o Pool Attention & 68.20 & 67.45 & 68.01 \\
w/o 1D Temporal Embedding & 66.33 & 64.10 & 65.44 \\
w/o 2D Spatial Embedding & 67.77 & 62.89 & 63.11 \\
w/o Peripheral Feature & 64.34 & 63.45 & 65.02 \\
w/o Fovea Mapping & 58.92 & 57.01 & 57.66 \\
\bottomrule
\end{tabular}
}
\end{table}

\section{Conclusion}
% We present a novel framework that use eye movement to predict diagnostic intent. 
% Limit: chưa đem qua các data khác được vì EGD REFLACX là cái duy nhất có pair nhiều class cho từng gaze point. 
% coco freeview thì không có class (cần check)
% cocosearch thì không có freeview. (cần check)
\textbf{Discussion.}
In this work, we introduce a novel paradigm for interpreting radiologists’ eye movements through the lens of intention, shifting the focus from mimicry-based modeling to a cognitively grounded understanding of visual search behavior. Our RadGazeIntent framework classifies each fixation point by its likely diagnostic purpose. We further propose three datasets, RadSeq, RadExplore, and RadHybrid, each representing a distinct observation of how radiologists allocate visual attention in practice. These settings allowed us to empirically test competing theories of gaze behavior in radiology, from structured checklist-based scanning to reactive exploration. Experiments show RadGazeIntent consistently outperforms existing baselines across all datasets. 

\textbf{Limitations.}
Despite these advances, several limitations remain. First, "intention" is inherently abstract and cannot be directly observed in eye tracking datasets, we infer intention from our observations on radiologist's behavior, which may not perfectly reflect cognitive processes. Second, while we model intention at the individual fixation level, not all fixations may map cleanly to a specific diagnostic objective (early or exploratory phases of scanning). This ambiguity introduces inherent noise that may confound classification. Finally, while conceptually diverse, our datasets focus exclusively on chest X-rays;; radiologists' behaviors in other imaging modalities, e.g., CT or MRI, may follow different patterns. Future work presents exciting opportunities to extend our work to other modalities, potentially revealing universal patterns of expert visual reasoning.

\textbf{Broader Impact.} By decoding the intent behind radiologists’ gaze behavior, our approach opens new pathways for developing interactive, intention-aware systems that can collaborate with rather than replace human experts. Potential applications include gaze-guided report generation, intention-aware assistance during training, and real-time feedback systems that adapt to a user's diagnostic focus. Furthermore, understanding gaze intent has implications beyond radiology. For example, in surgical navigation, pathology slide review, and even education platforms that teach visual diagnostic skills.

% \begin{acks}
% \end{acks}

\clearpage
\bibliographystyle{ACM-Reference-Format}
\bibliography{main}


\begin{thebibliography}{54}

%%% the Web address.

\ifx \showCODEN    \undefined \def \showCODEN     #1{\unskip}     \fi
\ifx \showISBNx    \undefined \def \showISBNx     #1{\unskip}     \fi
\ifx \showISBNxiii \undefined \def \showISBNxiii  #1{\unskip}     \fi
\ifx \showISSN     \undefined \def \showISSN      #1{\unskip}     \fi
\ifx \showLCCN     \undefined \def \showLCCN      #1{\unskip}     \fi
\ifx \shownote     \undefined \def \shownote      #1{#1}          \fi
\ifx \showarticletitle \undefined \def \showarticletitle #1{#1}   \fi
\ifx \showURL      \undefined \def \showURL       {\relax}        \fi
% invisible to TeX
\providecommand\bibfield[2]{#2}
\providecommand\bibinfo[2]{#2}
\providecommand\natexlab[1]{#1}
\providecommand\showeprint[2][]{arXiv:#2}

\bibitem[Aydemir et~al\mbox{.}(2023)]%
        {bahar:2023:tempsal}
\bibfield{author}{\bibinfo{person}{Bahar Aydemir}, \bibinfo{person}{Ludo Hoffstetter}, \bibinfo{person}{Tong Zhang}, \bibinfo{person}{Mathieu Salzmann}, {and} \bibinfo{person}{Sabine Susstrunk}.} \bibinfo{year}{2023}\natexlab{}.
\newblock \showarticletitle{{TempSAL} - Uncovering Temporal Information for Deep Saliency Prediction.}. In \bibinfo{booktitle}{\emph{Proceedings of the IEEE Conference on Computer Vision and Pattern Recognition (CVPR)}}.
\newblock


\bibitem[Azizi et~al\mbox{.}(2021)]%
        {azizi2021big}
\bibfield{author}{\bibinfo{person}{Shekoofeh Azizi}, \bibinfo{person}{Basil Mustafa}, \bibinfo{person}{Fiona Ryan}, \bibinfo{person}{Zachary Beaver}, \bibinfo{person}{Jan Freyberg}, \bibinfo{person}{Jonathan Deaton}, \bibinfo{person}{Aaron Loh}, \bibinfo{person}{Alan Karthikesalingam}, \bibinfo{person}{Simon Kornblith}, \bibinfo{person}{Ting Chen}, {et~al\mbox{.}}} \bibinfo{year}{2021}\natexlab{}.
\newblock \showarticletitle{Big self-supervised models advance medical image classification}. In \bibinfo{booktitle}{\emph{Proceedings of the IEEE/CVF International Conference on Computer Vision}}. \bibinfo{pages}{3478--3488}.
\newblock


\bibitem[Bertram et~al\mbox{.}(2013)]%
        {bertram2013effect}
\bibfield{author}{\bibinfo{person}{Raymond Bertram}, \bibinfo{person}{Laura Helle}, \bibinfo{person}{Johanna~K Kaakinen}, {and} \bibinfo{person}{Erkki Svedstr{\"o}m}.} \bibinfo{year}{2013}\natexlab{}.
\newblock \showarticletitle{The effect of expertise on eye movement behaviour in medical image perception}.
\newblock \bibinfo{journal}{\emph{PloS one}} \bibinfo{volume}{8}, \bibinfo{number}{6} (\bibinfo{year}{2013}), \bibinfo{pages}{e66169}.
\newblock


\bibitem[Bhattacharya et~al\mbox{.}(2022)]%
        {bhattacharya2022gazeradar}
\bibfield{author}{\bibinfo{person}{Moinak Bhattacharya}, \bibinfo{person}{Shubham Jain}, {and} \bibinfo{person}{Prateek Prasanna}.} \bibinfo{year}{2022}\natexlab{}.
\newblock \showarticletitle{Gazeradar: A gaze and radiomics-guided disease localization framework}. In \bibinfo{booktitle}{\emph{International Conference on Medical Image Computing and Computer-Assisted Intervention}}. Springer, \bibinfo{pages}{686--696}.
\newblock


\bibitem[Bigolin~Lanfredi et~al\mbox{.}(2022)]%
        {bigolin2022reflacx}
\bibfield{author}{\bibinfo{person}{Ricardo Bigolin~Lanfredi}, \bibinfo{person}{Mingyuan Zhang}, \bibinfo{person}{William~F Auffermann}, \bibinfo{person}{Jessica Chan}, \bibinfo{person}{Phuong-Anh~T Duong}, \bibinfo{person}{Vivek Srikumar}, \bibinfo{person}{Trafton Drew}, \bibinfo{person}{Joyce~D Schroeder}, {and} \bibinfo{person}{Tolga Tasdizen}.} \bibinfo{year}{2022}\natexlab{}.
\newblock \showarticletitle{REFLACX, a dataset of reports and eye-tracking data for localization of abnormalities in chest x-rays}.
\newblock \bibinfo{journal}{\emph{Scientific data}} \bibinfo{volume}{9}, \bibinfo{number}{1} (\bibinfo{year}{2022}), \bibinfo{pages}{350}.
\newblock


\bibitem[Chakraborty and otherss(2022)]%
        {souradeep:2022:agdf}
\bibfield{author}{\bibinfo{person}{Souradeep Chakraborty} {and} \bibinfo{person}{otherss}.} \bibinfo{year}{2022}\natexlab{}.
\newblock \showarticletitle{Predicting Visual Attention in Graphic Design Documents.}
\newblock \bibinfo{journal}{\emph{IEEE Transactions on Multimedia (TMM)}} (\bibinfo{year}{2022}).
\newblock


\bibitem[Chen et~al\mbox{.}(2020)]%
        {shi:2020:air}
\bibfield{author}{\bibinfo{person}{Shi Chen}, \bibinfo{person}{Ming Jiang}, \bibinfo{person}{Jinhui Yang}, {and} \bibinfo{person}{Qi Zhao}.} \bibinfo{year}{2020}\natexlab{}.
\newblock \showarticletitle{{AiR}: Attention with Reasoning Capability.}. In \bibinfo{booktitle}{\emph{Proceedings of the European Conference on Computer Vision (ECCV)}}.
\newblock


\bibitem[Chen et~al\mbox{.}(2023)]%
        {shi:2023:personalsaliency}
\bibfield{author}{\bibinfo{person}{Shi Chen}, \bibinfo{person}{Nachiappan Valliappan}, \bibinfo{person}{Shaolei Shen}, \bibinfo{person}{Xinyu Ye}, \bibinfo{person}{Kai Kohlhoff}, {and} \bibinfo{person}{Junfeng He}.} \bibinfo{year}{2023}\natexlab{}.
\newblock \showarticletitle{Learning from Unique Perspectives: User-aware Saliency Modeling.}. In \bibinfo{booktitle}{\emph{Proceedings of the IEEE Conference on Computer Vision and Pattern Recognition (CVPR)}}.
\newblock


\bibitem[Chen et~al\mbox{.}(2021)]%
        {xianyu:2021:vqa}
\bibfield{author}{\bibinfo{person}{Xianyu Chen}, \bibinfo{person}{Ming Jiang}, {and} \bibinfo{person}{Qi Zhao}.} \bibinfo{year}{2021}\natexlab{}.
\newblock \showarticletitle{Predicting Human Scanpaths in Visual Question Answering.}. In \bibinfo{booktitle}{\emph{Proceedings of the IEEE Conference on Computer Vision and Pattern Recognition (CVPR)}}.
\newblock


\bibitem[Chen et~al\mbox{.}(2024a)]%
        {xianyu:2024:individualscanpath}
\bibfield{author}{\bibinfo{person}{Xianyu Chen}, \bibinfo{person}{Ming Jiang}, {and} \bibinfo{person}{Qi Zhao}.} \bibinfo{year}{2024}\natexlab{a}.
\newblock \showarticletitle{Beyond Average: Individualized Visual Scanpath Prediction.}. In \bibinfo{booktitle}{\emph{Proceedings of the IEEE Conference on Computer Vision and Pattern Recognition (CVPR)}}.
\newblock


\bibitem[Chen et~al\mbox{.}(2024b)]%
        {chen2024isp}
\bibfield{author}{\bibinfo{person}{Xianyu Chen}, \bibinfo{person}{Ming Jiang}, {and} \bibinfo{person}{Qi Zhao}.} \bibinfo{year}{2024}\natexlab{b}.
\newblock \showarticletitle{Beyond Average: Individualized Visual Scanpath Prediction}. In \bibinfo{booktitle}{\emph{Proceedings of the IEEE/CVF Conference on Computer Vision and Pattern Recognition}}. \bibinfo{pages}{25420--25431}.
\newblock


\bibitem[Chen et~al\mbox{.}(2022)]%
        {chen2022cocofreeview1}
\bibfield{author}{\bibinfo{person}{Yupei Chen} {et~al\mbox{.}}} \bibinfo{year}{2022}\natexlab{}.
\newblock \showarticletitle{Characterizing Target-Absent Human Attention}. In \bibinfo{booktitle}{\emph{Proceedings of the IEEE/CVF Conference on Computer Vision and Pattern Recognition Workshops}}. \bibinfo{pages}{5031--5040}.
\newblock


\bibitem[Chen and Sun(2018)]%
        {zhenzhong:2018:iorroi-lstm}
\bibfield{author}{\bibinfo{person}{Zhenzhong Chen} {and} \bibinfo{person}{Wanjie Sun}.} \bibinfo{year}{2018}\natexlab{}.
\newblock \showarticletitle{Scanpath Prediction for Visual Attention using {IOR-ROI LSTM}.}. In \bibinfo{booktitle}{\emph{International Joint Conference on Artificial Intelligence (IJCAI)}}.
\newblock


\bibitem[Cheng et~al\mbox{.}(2022)]%
        {cheng2022masked}
\bibfield{author}{\bibinfo{person}{Bowen Cheng}, \bibinfo{person}{Ishan Misra}, \bibinfo{person}{Alexander~G Schwing}, \bibinfo{person}{Alexander Kirillov}, {and} \bibinfo{person}{Rohit Girdhar}.} \bibinfo{year}{2022}\natexlab{}.
\newblock \showarticletitle{Masked-attention mask transformer for universal image segmentation}. In \bibinfo{booktitle}{\emph{Proceedings of the IEEE/CVF conference on computer vision and pattern recognition}}. \bibinfo{pages}{1290--1299}.
\newblock


\bibitem[Cornia et~al\mbox{.}(2018)]%
        {marcella:2018:sam}
\bibfield{author}{\bibinfo{person}{Marcella Cornia}, \bibinfo{person}{Lorenzo Baraldi}, \bibinfo{person}{Giuseppe Serra}, {and} \bibinfo{person}{Rita Cucchiara}.} \bibinfo{year}{2018}\natexlab{}.
\newblock \showarticletitle{Predicting Human Eye Fixations via an LSTM-based Saliency Attentive Model}.
\newblock \bibinfo{journal}{\emph{IEEE Transactions on Image Processing (IEEE TIP)}} (\bibinfo{year}{2018}).
\newblock


\bibitem[Fan et~al\mbox{.}(2021)]%
        {fan2021multiscale}
\bibfield{author}{\bibinfo{person}{Haoqi Fan}, \bibinfo{person}{Bo Xiong}, \bibinfo{person}{Karttikeya Mangalam}, \bibinfo{person}{Yanghao Li}, \bibinfo{person}{Zhicheng Yan}, \bibinfo{person}{Jitendra Malik}, {and} \bibinfo{person}{Christoph Feichtenhofer}.} \bibinfo{year}{2021}\natexlab{}.
\newblock \showarticletitle{Multiscale vision transformers}. In \bibinfo{booktitle}{\emph{Proceedings of the IEEE/CVF international conference on computer vision}}. \bibinfo{pages}{6824--6835}.
\newblock


\bibitem[Fosco et~al\mbox{.}(2020)]%
        {camilo:2020:umsi}
\bibfield{author}{\bibinfo{person}{Camilo Fosco}, \bibinfo{person}{Vincent Casser}, \bibinfo{person}{Amish~Kumar Bedi}, \bibinfo{person}{Peter O'Donovan}, \bibinfo{person}{Aaron Hertzmann}, {and} \bibinfo{person}{Zoya Bylinskii}.} \bibinfo{year}{2020}\natexlab{}.
\newblock \showarticletitle{Predicting Visual Importance Across Graphic Design Types.}. In \bibinfo{booktitle}{\emph{ACM Symposium on User Interface Software and Technology}}.
\newblock


\bibitem[Gazda et~al\mbox{.}(2021)]%
        {gazda2021self}
\bibfield{author}{\bibinfo{person}{Matej Gazda}, \bibinfo{person}{J{\'a}n Plavka}, \bibinfo{person}{Jakub Gazda}, {and} \bibinfo{person}{Peter Drotar}.} \bibinfo{year}{2021}\natexlab{}.
\newblock \showarticletitle{Self-supervised deep convolutional neural network for chest x-ray classification}.
\newblock \bibinfo{journal}{\emph{IEEE Access}}  \bibinfo{volume}{9} (\bibinfo{year}{2021}), \bibinfo{pages}{151972--151982}.
\newblock


\bibitem[He et~al\mbox{.}(2015)]%
        {he2015resnet}
\bibfield{author}{\bibinfo{person}{Kaiming He}, \bibinfo{person}{Xiangyu Zhang}, \bibinfo{person}{Shaoqing Ren}, {and} \bibinfo{person}{Jian Sun}.} \bibinfo{year}{2015}\natexlab{}.
\newblock \showarticletitle{Deep residual learning for image recognition. arXiv e-prints}.
\newblock \bibinfo{journal}{\emph{arXiv preprint arXiv:1512.03385}}  \bibinfo{volume}{10} (\bibinfo{year}{2015}).
\newblock


\bibitem[Hsieh et~al\mbox{.}(2024)]%
        {hsieh2024eyexnet}
\bibfield{author}{\bibinfo{person}{Chihcheng Hsieh}, \bibinfo{person}{Andr{\'e} Lu{\'\i}s}, \bibinfo{person}{Jos{\'e} Neves}, \bibinfo{person}{Isabel~Blanco Nobre}, \bibinfo{person}{Sandra~Costa Sousa}, \bibinfo{person}{Chun Ouyang}, \bibinfo{person}{Joaquim Jorge}, {and} \bibinfo{person}{Catarina Moreira}.} \bibinfo{year}{2024}\natexlab{}.
\newblock \showarticletitle{EyeXNet: Enhancing Abnormality Detection and Diagnosis via Eye-Tracking and X-ray Fusion}.
\newblock \bibinfo{journal}{\emph{Machine Learning and Knowledge Extraction}} \bibinfo{volume}{6}, \bibinfo{number}{2} (\bibinfo{year}{2024}), \bibinfo{pages}{1055--1071}.
\newblock


\bibitem[Huang et~al\mbox{.}(2017)]%
        {huang2017densenet}
\bibfield{author}{\bibinfo{person}{Gao Huang}, \bibinfo{person}{Zhuang Liu}, \bibinfo{person}{Laurens Van Der~Maaten}, {and} \bibinfo{person}{Kilian~Q Weinberger}.} \bibinfo{year}{2017}\natexlab{}.
\newblock \showarticletitle{Densely connected convolutional networks}. In \bibinfo{booktitle}{\emph{Proceedings of the IEEE conference on computer vision and pattern recognition}}. \bibinfo{pages}{4700--4708}.
\newblock


\bibitem[Huang et~al\mbox{.}(2015)]%
        {xun:2015:salicon}
\bibfield{author}{\bibinfo{person}{Xun Huang}, \bibinfo{person}{Chengyao Shen}, \bibinfo{person}{Xavier Boix}, {and} \bibinfo{person}{Qi Zhao}.} \bibinfo{year}{2015}\natexlab{}.
\newblock \showarticletitle{{SALICON}: Reducing the Semantic Gap in Saliency Prediction by Adapting Deep Neural Networks}. In \bibinfo{booktitle}{\emph{Proceedings of the IEEE International Conference on Computer Vision (ICCV)}}.
\newblock


\bibitem[Irvin et~al\mbox{.}(2019)]%
        {irvin2019chexpert}
\bibfield{author}{\bibinfo{person}{Jeremy Irvin} {et~al\mbox{.}}} \bibinfo{year}{2019}\natexlab{}.
\newblock \showarticletitle{Chexpert: A large chest radiograph dataset with uncertainty labels and expert comparison}. In \bibinfo{booktitle}{\emph{Proceedings of the AAAI conference on artificial intelligence}}, Vol.~\bibinfo{volume}{33}. \bibinfo{pages}{590--597}.
\newblock


\bibitem[Jia and Bruce(2020)]%
        {sen:2020:eml}
\bibfield{author}{\bibinfo{person}{Sen Jia} {and} \bibinfo{person}{Neil D.~B. Bruce}.} \bibinfo{year}{2020}\natexlab{}.
\newblock \showarticletitle{{EML-NET}:An Expandable Multi-Layer NETwork for Saliency Prediction.}
\newblock \bibinfo{journal}{\emph{Image and Vision Computing}} (\bibinfo{year}{2020}).
\newblock


\bibitem[Karargyris et~al\mbox{.}(2021)]%
        {karargyris2021creation}
\bibfield{author}{\bibinfo{person}{Alexandros Karargyris} {et~al\mbox{.}}} \bibinfo{year}{2021}\natexlab{}.
\newblock \showarticletitle{Creation and validation of a chest X-ray dataset with eye-tracking and report dictation for AI development}.
\newblock \bibinfo{journal}{\emph{Scientific Data}} \bibinfo{volume}{8}, \bibinfo{number}{1} (\bibinfo{year}{2021}), \bibinfo{pages}{1--18}.
\newblock


\bibitem[Kümmerer et~al\mbox{.}(2022)]%
        {matthias:2022:deepgaze}
\bibfield{author}{\bibinfo{person}{Matthias Kümmerer}, \bibinfo{person}{Matthias Bethge}, {and} \bibinfo{person}{Thomas S.~A. Wallis}.} \bibinfo{year}{2022}\natexlab{}.
\newblock \showarticletitle{{DeepGaze III}: Modeling free-viewing human scanpaths with deep learning.}
\newblock \bibinfo{journal}{\emph{Journal of Vision (JoV)}} (\bibinfo{year}{2022}).
\newblock


\bibitem[Kümmerer et~al\mbox{.}(2016)]%
        {matthias:2016:deepgaze}
\bibfield{author}{\bibinfo{person}{Matthias Kümmerer}, \bibinfo{person}{Thomas S.~A. Wallis}, {and} \bibinfo{person}{Matthias Bethge}.} \bibinfo{year}{2016}\natexlab{}.
\newblock \showarticletitle{{DeepGaze II}: Reading fixations from deep features trained on object recognition.}
\newblock \bibinfo{journal}{\emph{arXiv preprint arXiv:1610.01563}} (\bibinfo{year}{2016}).
\newblock


\bibitem[Li et~al\mbox{.}(2023)]%
        {peizhao:2023:uniar}
\bibfield{author}{\bibinfo{person}{Peizhao Li}, \bibinfo{person}{Junfeng He}, \bibinfo{person}{Gang Li}, \bibinfo{person}{Rachit Bhargava}, \bibinfo{person}{Shaolei Shen}, \bibinfo{person}{Nachiappan Valliappan}, \bibinfo{person}{Youwei Liang}, \bibinfo{person}{Hongxiang Gu}, \bibinfo{person}{Venky Ramachandran}, \bibinfo{person}{Golnaz Farhadi}, \bibinfo{person}{Yang Li}, \bibinfo{person}{Kai~J Kohlhoff}, {and} \bibinfo{person}{Vidhya Navalpakkam}.} \bibinfo{year}{2023}\natexlab{}.
\newblock \showarticletitle{{UniAR}: Unifying Human Attention and Response Prediction on Visual Content.}
\newblock \bibinfo{journal}{\emph{arXiv preprint arXiv:2312.10175}} (\bibinfo{year}{2023}).
\newblock


\bibitem[Li et~al\mbox{.}(2018)]%
        {li2018thoracic}
\bibfield{author}{\bibinfo{person}{Zhe Li}, \bibinfo{person}{Chong Wang}, \bibinfo{person}{Mei Han}, \bibinfo{person}{Yuan Xue}, \bibinfo{person}{Wei Wei}, \bibinfo{person}{Li-Jia Li}, {and} \bibinfo{person}{Li Fei-Fei}.} \bibinfo{year}{2018}\natexlab{}.
\newblock \showarticletitle{Thoracic disease identification and localization with limited supervision}. In \bibinfo{booktitle}{\emph{Proceedings of the IEEE conference on computer vision and pattern recognition}}. \bibinfo{pages}{8290--8299}.
\newblock


\bibitem[Lin et~al\mbox{.}(2017)]%
        {lin2017fpn}
\bibfield{author}{\bibinfo{person}{Tsung-Yi Lin}, \bibinfo{person}{Piotr Doll{\'a}r}, \bibinfo{person}{Ross Girshick}, \bibinfo{person}{Kaiming He}, \bibinfo{person}{Bharath Hariharan}, {and} \bibinfo{person}{Serge Belongie}.} \bibinfo{year}{2017}\natexlab{}.
\newblock \showarticletitle{Feature pyramid networks for object detection}. In \bibinfo{booktitle}{\emph{Proceedings of the IEEE conference on computer vision and pattern recognition}}. \bibinfo{pages}{2117--2125}.
\newblock


\bibitem[Liu et~al\mbox{.}(2021)]%
        {liu2021self}
\bibfield{author}{\bibinfo{person}{Fengbei Liu}, \bibinfo{person}{Yu Tian}, \bibinfo{person}{Filipe~R Cordeiro}, \bibinfo{person}{Vasileios Belagiannis}, \bibinfo{person}{Ian Reid}, {and} \bibinfo{person}{Gustavo Carneiro}.} \bibinfo{year}{2021}\natexlab{}.
\newblock \showarticletitle{Self-supervised mean teacher for semi-supervised chest x-ray classification}. In \bibinfo{booktitle}{\emph{Machine Learning in Medical Imaging: 12th International Workshop, MLMI 2021, Held in Conjunction with MICCAI 2021, Strasbourg, France, September 27, 2021, Proceedings 12}}. Springer, \bibinfo{pages}{426--436}.
\newblock


\bibitem[Liu et~al\mbox{.}(2019)]%
        {liu2019align}
\bibfield{author}{\bibinfo{person}{Jingyu Liu}, \bibinfo{person}{Gangming Zhao}, \bibinfo{person}{Yu Fei}, \bibinfo{person}{Ming Zhang}, \bibinfo{person}{Yizhou Wang}, {and} \bibinfo{person}{Yizhou Yu}.} \bibinfo{year}{2019}\natexlab{}.
\newblock \showarticletitle{Align, attend and locate: Chest x-ray diagnosis via contrast induced attention network with limited supervision}. In \bibinfo{booktitle}{\emph{Proceedings of the IEEE/CVF International Conference on Computer Vision}}. \bibinfo{pages}{10632--10641}.
\newblock


\bibitem[Liu et~al\mbox{.}(2020)]%
        {liu2020semi}
\bibfield{author}{\bibinfo{person}{Quande Liu}, \bibinfo{person}{Lequan Yu}, \bibinfo{person}{Luyang Luo}, \bibinfo{person}{Qi Dou}, {and} \bibinfo{person}{Pheng~Ann Heng}.} \bibinfo{year}{2020}\natexlab{}.
\newblock \showarticletitle{Semi-supervised medical image classification with relation-driven self-ensembling model}.
\newblock \bibinfo{journal}{\emph{IEEE transactions on medical imaging}} \bibinfo{volume}{39}, \bibinfo{number}{11} (\bibinfo{year}{2020}), \bibinfo{pages}{3429--3440}.
\newblock


\bibitem[Loshchilov(2017)]%
        {loshchilov2017adamw}
\bibfield{author}{\bibinfo{person}{I Loshchilov}.} \bibinfo{year}{2017}\natexlab{}.
\newblock \showarticletitle{Decoupled weight decay regularization}.
\newblock \bibinfo{journal}{\emph{arXiv preprint arXiv:1711.05101}} (\bibinfo{year}{2017}).
\newblock


\bibitem[Mondal et~al\mbox{.}(2023)]%
        {sounak:2023:gazeformer}
\bibfield{author}{\bibinfo{person}{Sounak Mondal} {et~al\mbox{.}}} \bibinfo{year}{2023}\natexlab{}.
\newblock \showarticletitle{{Gazeformer}: Scalable, Effective and Fast Prediction of Goal-Directed Human Attention.}. In \bibinfo{booktitle}{\emph{Proceedings of the IEEE Conference on Computer Vision and Pattern Recognition (CVPR)}}.
\newblock


\bibitem[Neves et~al\mbox{.}(2024)]%
        {neves2024shedding}
\bibfield{author}{\bibinfo{person}{Jos{\'e} Neves}, \bibinfo{person}{Chihcheng Hsieh}, \bibinfo{person}{Isabel~Blanco Nobre}, \bibinfo{person}{Sandra~Costa Sousa}, \bibinfo{person}{Chun Ouyang}, \bibinfo{person}{Anderson Maciel}, \bibinfo{person}{Andrew Duchowski}, \bibinfo{person}{Joaquim Jorge}, {and} \bibinfo{person}{Catarina Moreira}.} \bibinfo{year}{2024}\natexlab{}.
\newblock \showarticletitle{Shedding light on ai in radiology: A systematic review and taxonomy of eye gaze-driven interpretability in deep learning}.
\newblock \bibinfo{journal}{\emph{European Journal of Radiology}} (\bibinfo{year}{2024}), \bibinfo{pages}{111341}.
\newblock


\bibitem[Peng et~al\mbox{.}(2024)]%
        {peng2024eye}
\bibfield{author}{\bibinfo{person}{Peixi Peng}, \bibinfo{person}{Wanshu Fan}, \bibinfo{person}{Yue Shen}, \bibinfo{person}{Wenfei Liu}, \bibinfo{person}{Xin Yang}, \bibinfo{person}{Qiang Zhang}, \bibinfo{person}{Xiaopeng Wei}, {and} \bibinfo{person}{Dongsheng Zhou}.} \bibinfo{year}{2024}\natexlab{}.
\newblock \showarticletitle{Eye gaze guided cross-modal alignment network for radiology report generation}.
\newblock \bibinfo{journal}{\emph{IEEE Journal of Biomedical and Health Informatics}} (\bibinfo{year}{2024}).
\newblock


\bibitem[Pham et~al\mbox{.}(2024a)]%
        {pham2024ai}
\bibfield{author}{\bibinfo{person}{Trong~Thang Pham}, \bibinfo{person}{Jacob Brecheisen}, \bibinfo{person}{Anh Nguyen}, \bibinfo{person}{Hien Nguyen}, {and} \bibinfo{person}{Ngan Le}.} \bibinfo{year}{2024}\natexlab{a}.
\newblock \showarticletitle{I-AI: A Controllable \& Interpretable AI System for Decoding Radiologists' Intense Focus for Accurate CXR Diagnoses}. In \bibinfo{booktitle}{\emph{Proceedings of the IEEE/CVF Winter Conference on Applications of Computer Vision}}. \bibinfo{pages}{7850--7859}.
\newblock


\bibitem[Pham et~al\mbox{.}(2024b)]%
        {pham2024gazesearch}
\bibfield{author}{\bibinfo{person}{Trong~Thang Pham}, \bibinfo{person}{Tien-Phat Nguyen}, \bibinfo{person}{Yuki Ikebe}, \bibinfo{person}{Akash Awasthi}, \bibinfo{person}{Zhigang Deng}, \bibinfo{person}{Carol~C Wu}, \bibinfo{person}{Hien Nguyen}, {and} \bibinfo{person}{Ngan Le}.} \bibinfo{year}{2024}\natexlab{b}.
\newblock \showarticletitle{GazeSearch: Radiology Findings Search Benchmark}.
\newblock \bibinfo{journal}{\emph{arXiv preprint arXiv:2411.05780}} (\bibinfo{year}{2024}).
\newblock


\bibitem[Qiu et~al\mbox{.}(2023)]%
        {mengyu:2023:scanpath}
\bibfield{author}{\bibinfo{person}{Mengyu Qiu}, \bibinfo{person}{Yi Guo}, \bibinfo{person}{Mingguang Zhang}, \bibinfo{person}{Jingwei Zhang}, \bibinfo{person}{Tian Lan}, {and} \bibinfo{person}{Zhilin Liu}.} \bibinfo{year}{2023}\natexlab{}.
\newblock \showarticletitle{Simulating Human Visual System Based on Vision Transformer.}. In \bibinfo{booktitle}{\emph{Proceedings of the 2023 ACM Symposium on Spatial User Interaction}}.
\newblock


\bibitem[Rajpurkar et~al\mbox{.}(2017)]%
        {rajpurkar2017chexnet}
\bibfield{author}{\bibinfo{person}{Pranav Rajpurkar}, \bibinfo{person}{Jeremy Irvin}, \bibinfo{person}{Kaylie Zhu}, \bibinfo{person}{Brandon Yang}, \bibinfo{person}{Hershel Mehta}, \bibinfo{person}{Tony Duan}, \bibinfo{person}{Daisy Ding}, \bibinfo{person}{Aarti Bagul}, \bibinfo{person}{Curtis Langlotz}, \bibinfo{person}{Katie Shpanskaya}, {et~al\mbox{.}}} \bibinfo{year}{2017}\natexlab{}.
\newblock \showarticletitle{Chexnet: Radiologist-level pneumonia detection on chest x-rays with deep learning}.
\newblock \bibinfo{journal}{\emph{arXiv preprint arXiv:1711.05225}} (\bibinfo{year}{2017}).
\newblock


\bibitem[Smit et~al\mbox{.}(2020)]%
        {smit2020chexbert}
\bibfield{author}{\bibinfo{person}{Akshay Smit}, \bibinfo{person}{Saahil Jain}, \bibinfo{person}{Pranav Rajpurkar}, \bibinfo{person}{Anuj Pareek}, \bibinfo{person}{Andrew~Y. Ng}, {and} \bibinfo{person}{Matthew~P. Lungren}.} \bibinfo{year}{2020}\natexlab{}.
\newblock \bibinfo{title}{CheXbert: Combining Automatic Labelers and Expert Annotations for Accurate Radiology Report Labeling Using BERT}.
\newblock
\showeprint[arxiv]{2004.09167}~[cs.CL]


\bibitem[Sun et~al\mbox{.}(2019)]%
        {wanjie:2019:iorroi}
\bibfield{author}{\bibinfo{person}{Wanjie Sun}, \bibinfo{person}{Zhenzhong Chen}, {and} \bibinfo{person}{Feng Wu}.} \bibinfo{year}{2019}\natexlab{}.
\newblock \showarticletitle{Visual Scanpath Prediction using {IOR-ROI} Recurrent Mixture Density Network.}
\newblock \bibinfo{journal}{\emph{IEEE Transactions on Pattern Analysis and Machine Intelligence (IEEE TPAMI)}} (\bibinfo{year}{2019}).
\newblock


\bibitem[Taslimi et~al\mbox{.}(2022)]%
        {taslimi2022swinchex}
\bibfield{author}{\bibinfo{person}{Sina Taslimi}, \bibinfo{person}{Soroush Taslimi}, \bibinfo{person}{Nima Fathi}, \bibinfo{person}{Mohammadreza Salehi}, {and} \bibinfo{person}{Mohammad~Hossein Rohban}.} \bibinfo{year}{2022}\natexlab{}.
\newblock \showarticletitle{SwinCheX: Multi-label classification on chest X-ray images with transformers}.
\newblock \bibinfo{journal}{\emph{arXiv preprint arXiv:2206.04246}} (\bibinfo{year}{2022}).
\newblock


\bibitem[van Sonsbeek et~al\mbox{.}(2023)]%
        {van2023probabilistic}
\bibfield{author}{\bibinfo{person}{Tom van Sonsbeek}, \bibinfo{person}{Xiantong Zhen}, \bibinfo{person}{Dwarikanath Mahapatra}, {and} \bibinfo{person}{Marcel Worring}.} \bibinfo{year}{2023}\natexlab{}.
\newblock \showarticletitle{Probabilistic Integration of Object Level Annotations in Chest X-ray Classification}. In \bibinfo{booktitle}{\emph{Proceedings of the IEEE/CVF Winter Conference on Applications of Computer Vision}}. \bibinfo{pages}{3630--3640}.
\newblock


\bibitem[Vaswani(2017)]%
        {vaswani2017attention}
\bibfield{author}{\bibinfo{person}{A Vaswani}.} \bibinfo{year}{2017}\natexlab{}.
\newblock \showarticletitle{Attention is all you need}.
\newblock \bibinfo{journal}{\emph{Advances in Neural Information Processing Systems}} (\bibinfo{year}{2017}).
\newblock


\bibitem[Wang et~al\mbox{.}(2022)]%
        {wang2022mgca}
\bibfield{author}{\bibinfo{person}{Fuying Wang}, \bibinfo{person}{Yuyin Zhou}, \bibinfo{person}{Shujun Wang}, \bibinfo{person}{Varut Vardhanabhuti}, {and} \bibinfo{person}{Lequan Yu}.} \bibinfo{year}{2022}\natexlab{}.
\newblock \showarticletitle{Multi-granularity cross-modal alignment for generalized medical visual representation learning}.
\newblock \bibinfo{journal}{\emph{Advances in Neural Information Processing Systems}}  \bibinfo{volume}{35} (\bibinfo{year}{2022}), \bibinfo{pages}{33536--33549}.
\newblock


\bibitem[Wu et~al\mbox{.}(2021)]%
        {wu2021chest}
\bibfield{author}{\bibinfo{person}{Joy~T Wu}, \bibinfo{person}{Nkechinyere~N Agu}, \bibinfo{person}{Ismini Lourentzou}, \bibinfo{person}{Arjun Sharma}, \bibinfo{person}{Joseph~A Paguio}, \bibinfo{person}{Jasper~S Yao}, \bibinfo{person}{Edward~C Dee}, \bibinfo{person}{William Mitchell}, \bibinfo{person}{Satyananda Kashyap}, \bibinfo{person}{Andrea Giovannini}, {et~al\mbox{.}}} \bibinfo{year}{2021}\natexlab{}.
\newblock \showarticletitle{Chest imagenome dataset for clinical reasoning}.
\newblock \bibinfo{journal}{\emph{arXiv preprint arXiv:2108.00316}} (\bibinfo{year}{2021}).
\newblock


\bibitem[Xu et~al\mbox{.}(2020)]%
        {xu2020layoutlm}
\bibfield{author}{\bibinfo{person}{Yiheng Xu}, \bibinfo{person}{Minghao Li}, \bibinfo{person}{Lei Cui}, \bibinfo{person}{Shaohan Huang}, \bibinfo{person}{Furu Wei}, {and} \bibinfo{person}{Ming Zhou}.} \bibinfo{year}{2020}\natexlab{}.
\newblock \showarticletitle{Layoutlm: Pre-training of text and layout for document image understanding}. In \bibinfo{booktitle}{\emph{Proceedings of the 26th ACM SIGKDD international conference on knowledge discovery \& data mining}}. \bibinfo{pages}{1192--1200}.
\newblock


\bibitem[Yan et~al\mbox{.}(2018)]%
        {yan2018weakly}
\bibfield{author}{\bibinfo{person}{Chaochao Yan}, \bibinfo{person}{Jiawen Yao}, \bibinfo{person}{Ruoyu Li}, \bibinfo{person}{Zheng Xu}, {and} \bibinfo{person}{Junzhou Huang}.} \bibinfo{year}{2018}\natexlab{}.
\newblock \showarticletitle{Weakly supervised deep learning for thoracic disease classification and localization on chest x-rays}. In \bibinfo{booktitle}{\emph{Proceedings of the 2018 ACM international conference on bioinformatics, computational biology, and health informatics}}. \bibinfo{pages}{103--110}.
\newblock


\bibitem[Yang et~al\mbox{.}(2020)]%
        {zhibo:2020:cocosearch}
\bibfield{author}{\bibinfo{person}{Zhibo Yang} {et~al\mbox{.}}} \bibinfo{year}{2020}\natexlab{}.
\newblock \showarticletitle{Predicting Goal-directed Human Attention Using Inverse Reinforcement Learning.}. In \bibinfo{booktitle}{\emph{Proceedings of the IEEE Conference on Computer Vision and Pattern Recognition (CVPR)}}.
\newblock


\bibitem[Yang et~al\mbox{.}(2022)]%
        {zhibo:2022:targetabsent}
\bibfield{author}{\bibinfo{person}{Zhibo Yang}, \bibinfo{person}{Sounak Mondal}, \bibinfo{person}{Seoyoung Ahn}, \bibinfo{person}{Gregory Zelinsky}, \bibinfo{person}{Minh Hoai}, {and} \bibinfo{person}{Dimitris Samaras}.} \bibinfo{year}{2022}\natexlab{}.
\newblock \showarticletitle{Target-absent Human Attention.}. In \bibinfo{booktitle}{\emph{Proceedings of the European Conference on Computer Vision (ECCV)}}.
\newblock


\bibitem[Yang et~al\mbox{.}(2023)]%
        {zhibo:2023:hat}
\bibfield{author}{\bibinfo{person}{Zhibo Yang}, \bibinfo{person}{Sounak Mondal}, \bibinfo{person}{Seoyoung Ahn}, \bibinfo{person}{Gregory Zelinsky}, \bibinfo{person}{Minh Hoai}, {and} \bibinfo{person}{Dimitris Samaras}.} \bibinfo{year}{2023}\natexlab{}.
\newblock \showarticletitle{Predicting Human Attention using Computational Attention}.
\newblock \bibinfo{journal}{\emph{arXiv preprint arXiv:2303.09383v2}} (\bibinfo{year}{2023}).
\newblock


\bibitem[Yao et~al\mbox{.}(2018)]%
        {yao2018weakly}
\bibfield{author}{\bibinfo{person}{Li Yao}, \bibinfo{person}{Jordan Prosky}, \bibinfo{person}{Eric Poblenz}, \bibinfo{person}{Ben Covington}, {and} \bibinfo{person}{Kevin Lyman}.} \bibinfo{year}{2018}\natexlab{}.
\newblock \showarticletitle{Weakly supervised medical diagnosis and localization from multiple resolutions}.
\newblock \bibinfo{journal}{\emph{arXiv preprint arXiv:1803.07703}} (\bibinfo{year}{2018}).
\newblock


\end{thebibliography}

% \appendix









\end{document}